\documentclass[final,3p,times,twocolumn]{elsarticle}
\usepackage{amssymb}
\usepackage{amsmath}
\usepackage{url}
\usepackage{natbib}
\usepackage{booktabs}
\usepackage{cleveref}
\usepackage{lineno}
\usepackage{xcolor,graphicx}
\usepackage{pifont} % in preamble
\newcommand{\cmark}{\ding{51}}
\newcommand{\xmark}{\ding{55}}
\usepackage{bbm}
\usepackage{booktabs} % For professional table spacing
\usepackage{caption}  % For bold figure labels (e.g., Figure 1)
\usepackage{xcolor}   % For colored text

\newcommand{\fixedheightimg}[1]{%
  \vbox to 4.2cm{\vfill\hbox{\includegraphics[width=\linewidth, height=4.2cm, keepaspectratio]{#1}}\vfill}%
}

\newcommand{\subfigheader}[2]{\textbf{(#1) #2}}
\newcommand{\caplabel}[1]{\textit{#1}}
\newcommand{\targetword}[1]{\textbf{#1}}
\newcommand{\metricfont}[1]{\texttt{\footnotesize #1}} % Monospace for numbers
\newcommand{\metriclabel}[1]{\metricfont{#1:}}
\newcommand{\metricval}[1]{\metricfont{#1}}
\newcommand{\highlight}[1]{\textcolor{red}{\textbf{\metricfont{#1}}}}

\journal{Pattern Recognition Letters}

\begin{document}

\begin{frontmatter}

\author{Jae Joong Lee\corref{cor1}\fnref{label2}}
\ead{lee2161@purdue.edu}

\title{Language-Guided Invariance Probing of Vision–Language Models}


% \author{Jae Joong Lee} %% Author name

\affiliation{organization={Department of Computer Science, Purdue University},%Department and Organization
            addressline={305 N University Street}, 
            city={West Lafayette},
            postcode={47907}, 
            state={Indiana},
            country={USA}}

%% Abstract
\begin{abstract}
Vision--language models (VLMs) achieve strong zero-shot performance, yet their robustness to controlled linguistic perturbations remains poorly characterized. We propose \emph{Language-Guided Invariance Probing} (LGIP), a benchmark that quantifies (i) invariance to meaning-preserving paraphrases and (ii) sensitivity to meaning-changing semantic flips in image--text matching. On 40k MS~COCO images (five captions each), we generate paraphrases and rule-based flips that modify object category, color, or count, and evaluate frozen encoders with an invariance error, a sensitivity gap, and a positive-rate metric.

Experiments on nine VLMs show that EVA02-CLIP and large OpenCLIP variants achieve a favorable invariance--sensitivity trade-off, whereas SigLIP and SigLIP2 exhibit substantially higher invariance error and can score flipped captions above human descriptions, particularly for object and color edits. These behaviors are largely obscured by standard retrieval metrics, highlighting LGIP as a lightweight, model-agnostic diagnostic of linguistic robustness beyond conventional accuracy.
\end{abstract}

% \includegraphics[width=2.0\linewidth]{data/graphical_abstract/graphical_abstract.pdf}
% % \input{fig/gabs}

% \begin{highlights}
% \item Introduce LGIP, a benchmark for probing linguistic robustness of VLMs.

% \item Generate paraphrases and semantic flips on MS COCO captions automatically.

% \item Show EVA02-CLIP and OpenCLIP achieve a favorable invariance sensitivity tradeoff.

% \item Reveal SigLIP models often prefer flipped captions above human ground truth.
% \end{highlights}

%% Keywords

\end{frontmatter}


\section{Introduction}

Vision--language models (VLMs) such as CLIP~\cite{radford2021clip}, OpenCLIP~\cite{cherti2023openclip}, EVA02-CLIP~\cite{sun2023evaclip}, and SigLIP~\cite{zhai2023siglip} support zero-shot recognition, retrieval, and a wide range of multimodal systems. By aligning images and texts in a shared embedding space, they achieve strong benchmark performance without task-specific fine-tuning. However, standard evaluations provide limited insight into a basic behavioral question: how does a VLM respond when the \emph{text phrasing} changes while the \emph{image} remains fixed?

We argue that two complementary properties matter for robust image--text alignment. First, \emph{linguistic invariance}: similarity should remain stable under meaning-preserving paraphrases. Second, \emph{semantic sensitivity}: similarity should decrease when the caption is edited to contradict salient visual attributes such as object category, color, or count. Existing benchmarks largely conflate these behaviors into aggregate accuracy or retrieval scores, making it difficult to diagnose whether a model is brittle to surface form, under-responsive to semantic conflicts, or selectively vulnerable to specific perturbation types.

To address this gap, we introduce \emph{Language-Guided Invariance Probing} (LGIP), a lightweight diagnostic protocol for VLMs. Using MS~COCO~\cite{lin2014coco} (40k images, five human captions each), LGIP automatically generates two families of textual perturbations per image--caption pair: (i) paraphrases that preserve semantics while varying style and framing, and (ii) semantic flips that modify a targeted attribute (object, color, or count). Given a frozen encoder, LGIP summarizes behavior with an \emph{invariance error} that measures similarity variation under paraphrasing, and \emph{semantic sensitivity} and \emph{positive rate} that measure how reliably original captions outrank flipped counterparts.

Across nine popular VLMs, LGIP reveals a clear separation that is not apparent in conventional zero-shot benchmarks. CLIP-family and large OpenCLIP models, particularly EVA02-CLIP, combine low paraphrase-induced variation with consistent rejection of attribute-flipped captions. In contrast, SigLIP-family models often exhibit substantially higher invariance error and can prefer flipped captions over human descriptions for attribute-level edits, despite strong zero-shot performance. These behaviors directly impact applications that depend on diverse prompts, captions, or instructions that are semantically equivalent but stylistically varied.

LGIP is model-agnostic and easy to deploy: it uses existing caption corpora and rule-based perturbations, requires no access to model internals, and applies uniformly across architectures and training recipes. Despite its simplicity, it exposes systematic, model-specific weaknesses in linguistic robustness and semantic grounding.

Our contributions are threefold:
\begin{itemize}
    \item We introduce \textbf{Language-Guided Invariance Probing} (LGIP), a diagnostic benchmark for VLM robustness to meaning-preserving paraphrases and meaning-changing semantic flips in image--text similarity space.
    \item We apply LGIP on MS~COCO by generating paraphrases and attribute-targeted flips (object, color, count), and define metrics that disentangle invariance error from semantic sensitivity and positive rate.
    \item We analyze nine widely used VLMs and show that EVA02-CLIP and large OpenCLIP variants achieve a favorable invariance--sensitivity trade-off, while SigLIP-family models exhibit attribute-level failures that standard benchmarks do not capture.
\end{itemize}

\section{Related Work}
\noindent \paragraph{\textbf{Vision--Language Pretraining}}
CLIP learns joint image--text representations via contrastive pretraining on web-scale image--caption pairs~\cite{radford2021clip}. OpenCLIP provides reproducible scaling studies and public models trained on LAION~\cite{cherti2023openclip}, while EVA-CLIP improves training recipes and backbones for stronger zero-shot transfer~\cite{sun2023evaclip}. SigLIP replaces softmax contrastive learning with a pairwise sigmoid objective that reduces dependence on global batch statistics and can improve training efficiency~\cite{zhai2023siglip}; SigLIP2 extends this line with additional pretraining signals and data curation~\cite{tschannen2025siglip2}.
Moreover, prompt-based weakly supervised vision--language pretraining has been explored to improve transfer with reduced reliance on fully paired supervision~\cite{guo2025prompt_wsvlp}.
We treat pretrained VLM encoders as fixed backbones and focus on diagnosing their linguistic robustness.
\noindent \paragraph{\textbf{Robustness and Evaluation of VLMs}}
Recent work highlights that standard zero-shot or retrieval benchmarks can miss systematic failure modes. MMRobust benchmarks multimodal robustness under distribution shift across tasks~\cite{qiu2022mmrobust}, and granularity-oriented benchmarks stress sensitivity to label specificity~\cite{xu2024granularity}. Winoground probes visio-linguistic compositionality with minimal lexical changes, where many models perform near chance~\cite{thrush2022winoground}. ViLP targets reliance on language priors using controlled out-of-distribution constructions~\cite{luo2025vilp}. In contrast, LGIP isolates \emph{caption-only} perturbations while holding the image fixed and reports continuous, interpretable metrics at the embedding-similarity level.
\noindent \paragraph{\textbf{Behavioral Testing via Text Perturbations}}
Our protocol is inspired by capability-based behavioral testing in NLP. CheckList emphasizes targeted, contrastive tests for invariance and directional sensitivity~\cite{ribeiro2020checklist}, and TextAttack provides modular perturbations for stress-testing models~\cite{morris2020textattack}. We adapt this perspective to VLM encoders by generating meaning-preserving paraphrases and meaning-changing semantic flips grounded in MS~COCO captions~\cite{lin2014coco}, explicitly separating paraphrase invariance from semantic sensitivity.
\noindent \paragraph{\textbf{Relation to Existing Robustness Benchmarks}}
\begin{table}[t]
\centering
\scriptsize
\setlength{\tabcolsep}{3pt}
\renewcommand{\arraystretch}{0.95}
\caption{Cross-benchmark comparison of robustness diagnostic capabilities.}
\resizebox{\columnwidth}{!}{%
\begin{tabular}{lccc}
\toprule
Capability & Winoground~\cite{thrush2022winoground} & ViLP~\cite{luo2025vilp} & LGIP (Ours) \\
\midrule
Pair correctness                 & \cmark & \cmark & \cmark \\
Factorized text perturbations    & \xmark & \cmark & \cmark \\
Paraphrase invariance            & \xmark & \xmark & \cmark \\
Semantic flip sensitivity        & \xmark & \xmark & \cmark \\
Fine-grained diagnostics         & \xmark & \xmark & \cmark \\
\bottomrule
\end{tabular}}
\label{tab:cross_benchmark}
\end{table}

LGIP is related to prior VLM robustness diagnostics such as Winoground~\cite{thrush2022winoground}
and ViLP~\cite{luo2025vilp}, but targets a different diagnostic granularity.
As summarized in Table~\ref{tab:cross_benchmark}, Winoground provides a binary stress test for
pair correctness under compositional swaps, and ViLP probes robustness under controlled settings
that disentangle language priors from visual evidence.
In contrast, LGIP is designed as a scalable, caption-only diagnostic for embedding-based VLMs:
it introduces controllable and factorized text perturbations and reports continuous, interpretable
metrics that separate paraphrase invariance from semantic-flip sensitivity, enabling more
fine-grained attribution of failure modes.
\section{Language-Guided Invariance Probing}
\label{sec:lgip}
\begin{figure*}[t]
\centering
\includegraphics[width=\linewidth]{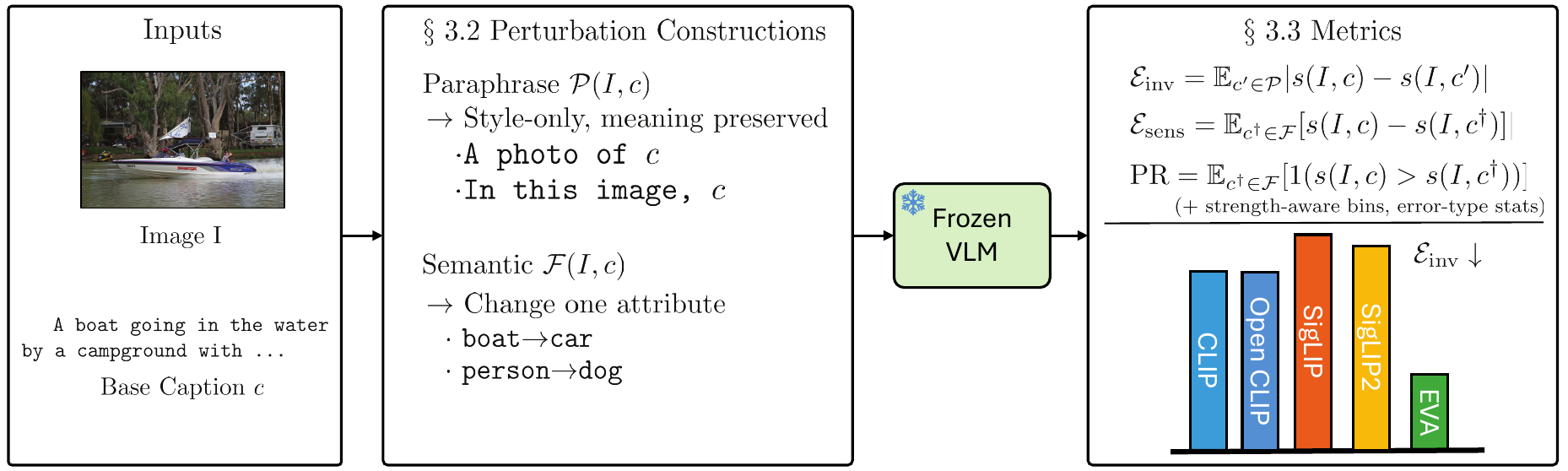}
\caption{Given an image and its human caption, we generate meaning-preserving paraphrases and semantic flips that change the object, color, or count, then feed all variants into a frozen vision language model to measure invariance and sensitivity of its similarity scores.}
\label{fig:teaser}
\end{figure*}

\subsection{Problem Setup}
We consider a vision--language model with an image encoder $f_{\text{img}}$ and a text encoder $f_{\text{text}}$, and define image--text similarity as cosine similarity between $\ell_2$-normalized embeddings:
\begin{equation}
s(I,c)=\operatorname{sim}\!\bigl(f_{\text{img}}(I),f_{\text{text}}(c)\bigr).
\end{equation}
Given a caption dataset $\mathcal{D}=\{(I_i,\{c_i^{(n)}\}_{n=1}^{N})\}_{i=1}^{M}$ (MS~COCO: $M{=}40{,}000$, $N{=}5$), LGIP constructs, for each base pair $(I,c)$, (i) a meaning-preserving \emph{paraphrase set} $\mathcal{P}(I,c)$ and (ii) a meaning-changing \emph{semantic flip set} $\mathcal{F}(I,c)$ that contradicts at least one salient attribute. LGIP probes two complementary behaviors of a fixed encoder: \emph{linguistic invariance} (stability under paraphrases) and \emph{semantic sensitivity} (whether contradictory captions are down-weighted), as illustrated in Fig.~\ref{fig:teaser}.

\subsection{Perturbation Constructions}
\label{subsec:perturbations}
\noindent \paragraph{\textbf{Filtering and constraints}}
All generated captions are (i) stripped and deduplicated by exact string match,
(ii) removed if identical to the original caption, and (iii) required to exceed a
minimum character length. We cap paraphrases and flips per caption by
$K_{\text{same}}$ and $K_{\text{diff}}$, respectively. For flips, we apply \emph{at most one}
lexical substitution per caption to isolate single-attribute changes.

\noindent \paragraph{\textbf{Token matching for flips}}
A flip is generated only when a caption contains a vocabulary token matched with
a word-boundary regular expression (e.g., \texttt{\textbackslash bdog\textbackslash b});
we replace the first match with a randomly sampled alternative from the same
vocabulary (excluding the original token) and label the flip by type
(\textit{object}, \textit{color}, \textit{number}). Consequently, flip coverage varies by
type depending on token occurrences in MS~COCO captions.

\noindent \paragraph{\textbf{Combined perturbations}}
To model realistic prompt variation, we additionally evaluate \emph{combined}
perturbations by composing paraphrase and flip operations: we first paraphrase
$c\!\rightarrow\!c'$ and then apply a semantic flip on $c'$ (with the flip type retained).

\noindent \paragraph{\textbf{Paraphrases}}
For each base caption $c$, we generate paraphrases by applying lightweight wrappers and prefixes (e.g., \texttt{``a photo of \{c\}''}, \texttt{``In this image, \{c\}''}), followed by deduplication and a per-caption cap $K_{\text{same}}$. To capture more realistic linguistic variation, we additionally generate \emph{advanced paraphrases} that introduce syntactic and lexical changes while preserving meaning, including passive-voice rewrites, synonym substitutions of non-critical modifiers, and minor structural edits (e.g., clause re-ordering). We stratify paraphrases into \textit{simple} and \textit{advanced} subsets and report invariance on each.

\noindent \paragraph{\textbf{Semantic flips}}
We generate semantic flips via a single targeted lexical substitution that changes one attribute at a time. Using small vocabularies of frequent MS~COCO nouns, color words, and number words, we replace a matched token with another token from the same list (regex match; one substitution per flip), remove duplicates, and cap the number of flips per caption at $K_{\text{diff}}$. Each flip is labeled by type, yielding $\mathcal{F}_{\text{obj}}(I,c)$, $\mathcal{F}_{\text{col}}(I,c)$, and $\mathcal{F}_{\text{num}}(I,c)$.

\noindent \paragraph{\textbf{Reproducibility}}
Full generation rules (template list, vocabularies, and deduplication) are provided in Appendix~\ref{app:generation_rules}.

\subsection{Metrics}
\label{subsec:metrics}
LGIP summarizes behavior with one invariance metric and two sensitivity metrics.

\noindent \paragraph{\textbf{Invariance error}}
For paraphrases $c'\!\in\!\mathcal{P}(I,c)$, invariance error measures similarity deviation under meaning-preserving edits:
\begin{equation}
\mathcal{E}_{\text{inv}}
=\mathbb{E}_{(I,c)}\,\mathbb{E}_{c'\in\mathcal{P}(I,c)}
\bigl[\lvert s(I,c)-s(I,c')\rvert\bigr].
\end{equation}
We also report $\mathcal{E}_{\text{inv}}^{\text{simple}}$ and $\mathcal{E}_{\text{inv}}^{\text{adv}}$ by restricting $\mathcal{P}$ to simple or advanced paraphrases.

\noindent \paragraph{\textbf{Semantic sensitivity and positive rate}}
For a flip $c^{\dagger}\!\in\!\mathcal{F}(I,c)$, define the gap
\begin{equation}
g(I,c,c^{\dagger}) = s(I,c)-s(I,c^{\dagger}).
\end{equation}
We report the mean gap
\begin{equation}
\mathcal{E}_{\text{sens}}
=\mathbb{E}_{(I,c)}\,\mathbb{E}_{c^{\dagger}\in\mathcal{F}(I,c)}\bigl[g(I,c,c^{\dagger})\bigr],
\end{equation}
and the positive rate
\begin{equation}
\text{PR}=\mathbb{E}_{(I,c)}\,\mathbb{E}_{c^{\dagger}\in\mathcal{F}(I,c)}
\bigl[\mathbb{I}\{s(I,c)>s(I,c^{\dagger})\}\bigr].
\end{equation}
PR $\approx 0.5$ corresponds to chance ordering; higher values indicate more reliable down-weighting of contradictory captions. All metrics can be computed per flip type $t\in\{\text{obj},\text{col},\text{num}\}$ by restricting $c^{\dagger}\in\mathcal{F}_t(I,c)$.

\section{Experiments}
\label{sec:experiments}

\subsection{Experimental Setup}

\noindent\textbf{Dataset.}
We evaluate LGIP on MS~COCO~\citep{lin2014coco} using $M{=}40{,}000$ training images and all $N{=}5$ human captions per image.
For each caption we generate up to $K_{\text{same}}{=}6$ paraphrases and $K_{\text{diff}}{=}6$ semantic flips following Section~\ref{subsec:perturbations}.
After filtering and deduplication, this yields $\sim$1.2M paraphrase comparisons and $80{,}632$ valid flip triples per model.

\noindent\textbf{Models.}
We test nine frozen dual-encoder VLMs: CLIP ViT-B/16 and ViT-L/14~\citep{radford2021clip}; OpenCLIP ViT-L/14 and ViT-H/14 (LAION-2B)~\citep{cherti2023openclip}; EVA02-CLIP L/14~\citep{sun2023evaclip}; SigLIP base (224, 384) and large (384)~\citep{zhai2023siglip}; and SigLIP2 base (224)~\citep{tschannen2025siglip2}.
LGIP perturbs only text while holding the image fixed.

\noindent\textbf{Protocol.}
We compute $\ell_2$-normalized image/text embeddings using each model’s official preprocessing and cosine similarity.
We report invariance error $\mathcal{E}_{\text{inv}}$, semantic sensitivity $\mathcal{E}_{\text{sens}}$, and positive rate (PR) as in Section~\ref{subsec:metrics}.
Unless stated otherwise, metrics aggregate all flip types; implementation details for generation (templates, vocabularies, deduplication) are in Appendix~\ref{app:generation_rules}.

\subsection{Main LGIP Results}
\label{subsec:main_results}
\begin{table}[t]
\centering
\small
\setlength{\tabcolsep}{4pt}
\caption{LGIP metrics on MS~COCO for nine vision--language models.
Lower invariance error $\mathcal{E}_{\text{inv}}$ is better.
Higher semantic sensitivity $\mathcal{E}_{\text{sens}}$ and positive rate (PR) are better.}
\begin{tabular}{lccc}
\toprule
Model & $\mathcal{E}_{\text{inv}} \downarrow$ & $\mathcal{E}_{\text{sens}} \uparrow$ & PR $\uparrow$\\
\midrule
CLIP ViT-B/16 (OpenAI)          & 0.008 &  0.024 & 0.866 \\
CLIP ViT-L/14 (OpenAI)          & 0.009 &  0.027 & 0.873 \\
OpenCLIP ViT-L/14 (LAION-2B)    & 0.008 &  0.046 & 0.898 \\
OpenCLIP ViT-H/14 (LAION-2B)    & 0.010 &  \textbf{0.050} & \textbf{0.908} \\
EVA02-CLIP L/14                 & \textbf{0.005} &  0.030 & 0.896 \\
\midrule
SigLIP base-p16-224             & 0.055 & -0.017 & 0.474 \\
SigLIP base-p16-384             & 0.058 & -0.021 & 0.464 \\
SigLIP large-p16-384            & 0.013 &  0.002 & 0.538 \\
SigLIP2 base-p16-224            & 0.041 &  0.008 & 0.649 \\
\bottomrule
\end{tabular}
\label{tab:main_lgip}
\end{table}

\begin{table}[t]
\centering
\small
\setlength{\tabcolsep}{4pt}
\caption{LGIP under \textbf{combined} perturbations (paraphrase + semantic flip).
Compared to Table~\ref{tab:main_lgip}, this setting is substantially harder,
revealing interaction effects between linguistic variation and semantic conflict.}
\begin{tabular}{lccc}
\toprule
Model & $\mathcal{E}_{\text{inv}} \downarrow$ & $\mathcal{E}_{\text{sens}}^{\text{comb}} \uparrow$ & PR$^{\text{comb}} \uparrow$\\
\midrule
CLIP ViT-B/16 (OpenAI)          & 0.008 &  0.019 & 0.747 \\
CLIP ViT-L/14 (OpenAI)          & 0.009 &  0.021 & 0.753 \\
OpenCLIP ViT-L/14 (LAION-2B)    & 0.009 &  0.044 & 0.865 \\
OpenCLIP ViT-H/14 (LAION-2B)    & 0.010 &  \textbf{0.049} & \textbf{0.882} \\
EVA02-CLIP L/14                 & \textbf{0.006} &  0.032 & 0.900 \\
\midrule
SigLIP base-p16-224             & 0.056 & -0.008 & 0.494 \\
SigLIP base-p16-384             & 0.058 & -0.008 & 0.505 \\
SigLIP large-p16-384            & 0.013 &  0.000 & 0.487 \\
SigLIP2 base-p16-224            & 0.039 &  0.013 & 0.637 \\
\bottomrule
\end{tabular}
\label{tab:lgip_combined}
\end{table}
\begin{figure}[t]
\centering
\includegraphics[width=\linewidth]{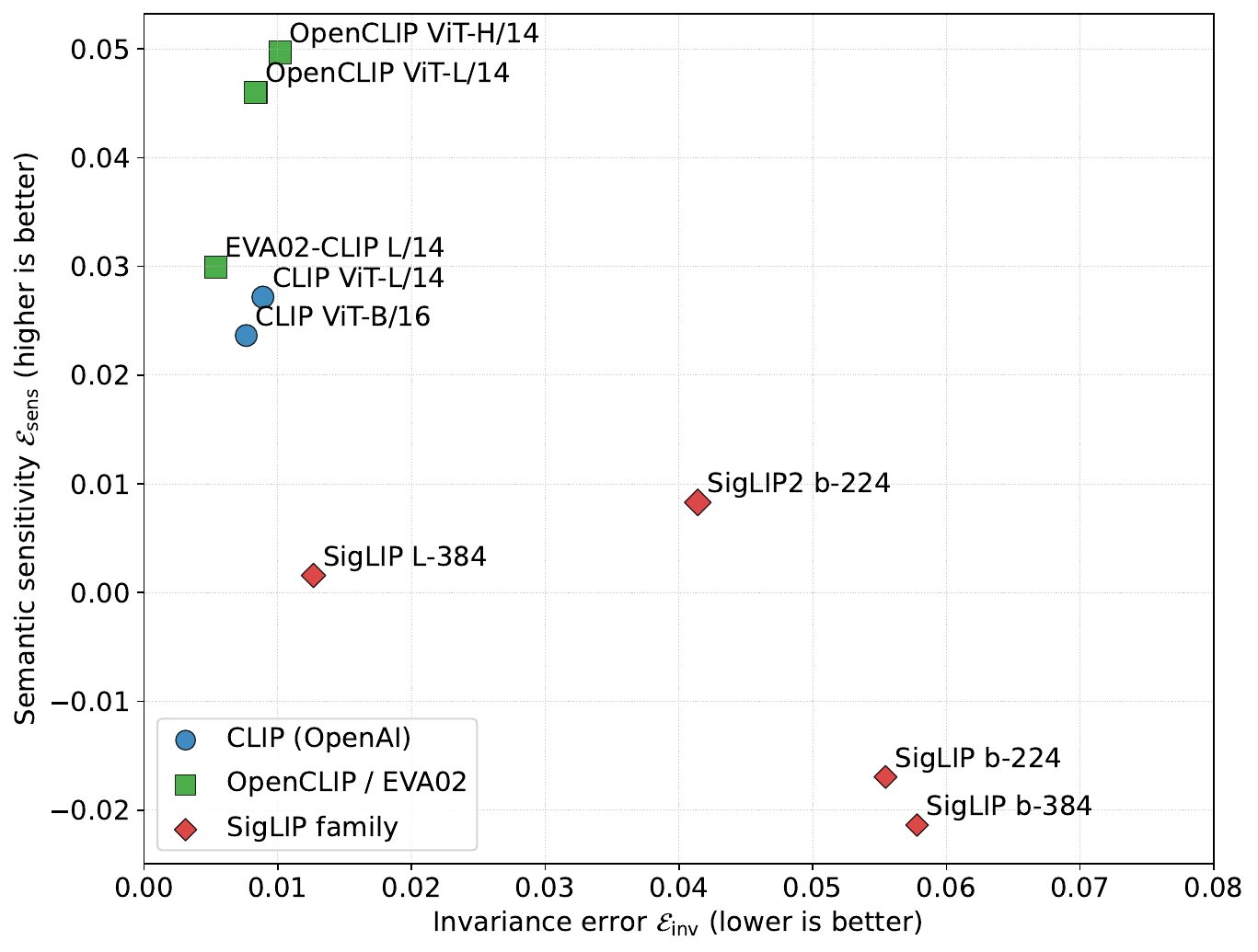}
\caption{Trade-off between linguistic invariance and semantic sensitivity on LGIP.
Each point corresponds to a model, plotted by invariance error ($\mathcal{E}_{\text{inv}}$, lower is better) and semantic sensitivity ($\mathcal{E}_{\text{sens}}$, higher is better).
EVA02-CLIP and large OpenCLIP models lie on a favorable frontier, while SigLIP-family models cluster in a region with high invariance error and low sensitivity.}
\label{fig:lgip_scatter}
\end{figure}
\begin{table}[t]
\centering
\small
\setlength{\tabcolsep}{4pt}
\caption{Paraphrase invariance under \textbf{simple} vs.\ \textbf{advanced} paraphrasing. ``Adv'' reports the relative increase from simple to advanced paraphrases.}
\begin{tabular}{lccc}
\toprule
Model & $\mathcal{E}_{\text{inv}}^{\text{simple}} \downarrow$ &
$\mathcal{E}_{\text{inv}}^{\text{adv}} \downarrow$ &
Adv$\uparrow$ (\%) \\
\midrule
CLIP ViT-B/16       & 0.007 & 0.009 & +14.7 \\
CLIP ViT-L/14         & 0.009 & 0.009 & +2.8  \\
OpenCLIP ViT-L/14   & 0.008 & 0.011 & +47.5 \\
OpenCLIP ViT-H/14  & 0.009 & 0.013 & +45.2 \\
EVA02-CLIP L/14               & \textbf{0.004} & 0.009 & +125.2 \\
\midrule
SigLIP base-p16-224           & 0.056 & 0.055 & -1.2  \\
SigLIP base-p16-384           & 0.058 & 0.058 & -0.8  \\
SigLIP large-p16-384          & 0.013 & 0.012 & -6.7  \\
SigLIP2 base-p16-224          & 0.041 & 0.035 & -15.4 \\
\bottomrule
\end{tabular}
\label{tab:paraphrase_difficulty}
\end{table}

Table~\ref{tab:main_lgip} reports LGIP metrics across all models, and Figure~\ref{fig:lgip_scatter} visualizes the invariance--sensitivity trade-off.
CLIP/OpenCLIP/EVA models occupy a favorable region, combining low paraphrase variance with strong separation between original captions and flips.
In contrast, SigLIP variants show substantially higher invariance error and near-chance ordering under flips (PR $\approx 0.5$), with SigLIP2 improving but still lagging CLIP-style models.
Table~\ref{tab:lgip_combined} further evaluates combined perturbations (paraphrase$+$flip), showing that the same split persists under composed edits.
Table~\ref{tab:paraphrase_difficulty} shows that advanced paraphrases increase $\mathcal{E}_{\text{inv}}$ across models, indicating that template robustness does not imply robustness to more realistic linguistic variation.

\subsection{Strength-Aware Semantic Flips}
\label{sec:strength}
\begin{table}[t]
\centering
\small
\setlength{\tabcolsep}{4pt}
\caption{Positive rate (PR) under strength-stratified semantic flips.
CLIP-family models respond monotonically to increasing perturbation strength,
whereas SigLIP exhibits weaker or inconsistent scaling.}
\resizebox{\linewidth}{!}{%
\begin{tabular}{lccc}
\toprule
Model & Weak PR & Medium PR & Strong PR \\
\midrule
CLIP ViT-B/16 (OpenAI)       & 0.55 & 0.86 & 0.96 \\
CLIP ViT-L/14 (OpenAI)       & 0.54 & 0.87 & 0.97 \\
OpenCLIP ViT-L/14 (LAION-2B) & 0.58 & 0.90 & 0.98 \\
OpenCLIP ViT-H/14 (LAION-2B) & 0.60 & 0.91 & 0.97 \\
EVA02-CLIP L/14              & 0.61 & 0.90 & 0.98 \\
\midrule
SigLIP base-p16-224          & 0.55 & 0.42 & 0.76 \\
SigLIP base-p16-384          & 0.57 & 0.41 & 0.77 \\
SigLIP large-p16-384         & 0.50 & 0.54 & 0.70 \\
SigLIP2 base-p16-224         & 0.51 & 0.68 & 0.12 \\
\bottomrule
\end{tabular}}
\label{tab:strength_bins}
\end{table}

Not all semantic flips induce the same degree of conflict.
We therefore assign each flip a continuous strength $s\in[0,1]$ by blending a type prior (color $<$ number $<$ object) with the semantic distance between the original and flipped captions in the model’s text-embedding space, and compute strength-weighted metrics.
We also stratify flips into weak/medium/strong bins using global quantiles.
Table~\ref{tab:strength_bins} reports PR by strength.
CLIP-style models show increasing PR with stronger conflicts, whereas SigLIP models exhibit flatter or less consistent scaling, revealing weaker sensitivity calibration.

\subsection{Error-Type Diagnostics}
\label{sec:error_type}
\begin{table}[t]
\centering
\small
\setlength{\tabcolsep}{4pt}
\caption{Error-type positive rate (PR) under semantic flips.
LGIP reveals attribute-specific failure modes that are invisible to aggregate metrics.}
\resizebox{\linewidth}{!}{%
\begin{tabular}{lccc}
\toprule
Model & Color PR$\uparrow$ & Number PR$\uparrow$ & Object PR$\uparrow$ \\
\midrule
CLIP ViT-B/16 (OpenAI)        & 0.83 & 0.66 & 0.95 \\
OpenCLIP ViT-H/14 (LAION-2B)  & \textbf{0.89} & \textbf{0.76} & \textbf{0.96} \\
EVA02-CLIP L/14               & 0.87 & 0.75 & 0.95 \\
\midrule
SigLIP base-p16-224           & 0.47 & 0.49 & 0.47 \\
SigLIP base-p16-384           & 0.47 & 0.48 & 0.46 \\
SigLIP2 base-p16-224          & 0.61 & 0.61 & 0.69 \\
\bottomrule
\end{tabular}}
\label{tab:error_type}
\end{table}

To localize failure modes, we compute metrics per flip type (\textit{object}, \textit{color}, \textit{number}).
Table~\ref{tab:error_type} shows that CLIP-family models maintain high PR across types, with the strongest separation on object-category flips.
SigLIP variants are markedly weaker and often uneven across attributes, while SigLIP2 improves on some types but remains substantially below CLIP on object-level conflicts.
These stratified results provide targeted diagnostics that are not visible from aggregate scores alone.

\subsection{Understanding the CLIP--SigLIP Gap}
Table~\ref{tab:main_lgip} indicates a large gap between CLIP/EVA and SigLIP under LGIP.
A key difference is the pretraining objective: CLIP-style models use a symmetric contrastive softmax loss that enforces batch-level relative ranking, whereas SigLIP optimizes a pairwise sigmoid loss that scores image--text pairs independently.
LGIP directly probes relative conflict resolution by contrasting an image with a human caption and its semantic flip, which aligns more naturally with the ranking pressure induced by contrastive softmax.
Consistent with this, CLIP-style models achieve high PR on object flips, while SigLIP remains near chance despite strong zero-shot classification performance.

\subsection{Flip-Type Breakdown}
\label{sec:fliptype}
\begin{table}[t]
\centering
\small
\setlength{\tabcolsep}{3pt}
\caption{Flip-type analysis on LGIP. Semantic sensitivity $\mathcal{E}_{\text{sens}}$ and positive rate (PR) are reported separately for object (Obj), color (Col), and count (Cnt) flips.}
\resizebox{\linewidth}{!}{%
\begin{tabular}{lcccccc}
\toprule
 & \multicolumn{2}{c}{Obj} & \multicolumn{2}{c}{Col} & \multicolumn{2}{c}{Cnt} \\
Model & Gap & PR & Gap & PR & Gap & PR \\
\midrule
CLIP ViT-B/16 (OpenAI) &
  0.037 & 0.955 &
  0.015 & 0.827 &
  0.004 & 0.665 \\

EVA02-CLIP L/14 &
  0.043 & 0.956 &
  0.021 & 0.869 &
  0.010 & 0.765 \\

SigLIP base-p16-224 &
 -0.017 & 0.466 &
 -0.017 & 0.477 &
 -0.016 & 0.493 \\

SigLIP2 base-p16-224 &
  0.011 & 0.695 &
  0.006 & 0.608 &
  0.006 & 0.607 \\
\bottomrule
\end{tabular}%
}
\label{tab:fliptype}
\end{table}

We further decompose results by flip type using the labels from Section~\ref{subsec:perturbations}.
CLIP/EVA models show positive gaps and high PR across object, color, and number edits, while SigLIP models remain near chance across types; SigLIP2 improves but still trails, especially on object flips.

\subsection{Qualitative Examples}
\label{subsec:qualitative}

\begin{figure}[t]
\centering
% Reduce column separation slightly for a tighter, cleaner look
\setlength{\tabcolsep}{2pt}

\begin{tabular}{cc}
% ==========================
% Row 1
% ==========================

% ----- Subfigure (a) -----
\begin{minipage}[t]{0.48\linewidth}
    \centering
    \fixedheightimg{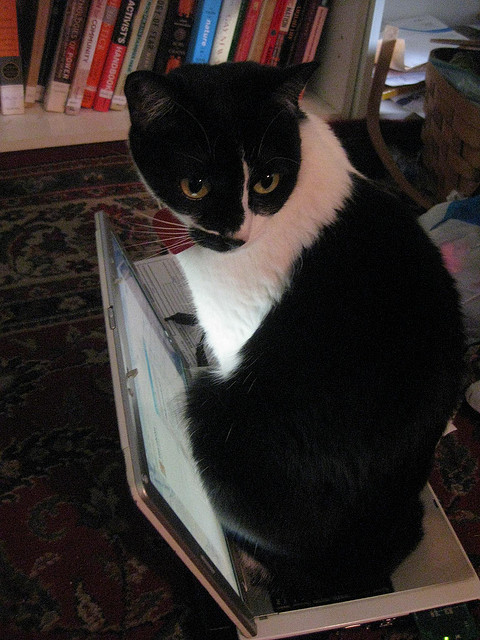}
    % \vspace{4pt}
    {\scriptsize
    \subfigheader{a}{cat $\rightarrow$ person}\\[2pt]
    \begin{tabular}{@{}l@{\hspace{0.7em}}p{0.82\linewidth}@{}}
        \caplabel{ORIG} & A cat sits on top of a computer.\\
                        & \metriclabel{E} \metricval{0.295},\; \metriclabel{S} \metricval{-0.116} \\ \addlinespace[2pt]
        \caplabel{FLIP} & A \targetword{person} sits on top of a computer.\\
                        & \metriclabel{E} \metricval{0.234},\; \metriclabel{S} \highlight{-0.042}
    \end{tabular}
    }
\end{minipage}
&
% ----- Subfigure (b) -----
\begin{minipage}[t]{0.48\linewidth}
    \centering
    \fixedheightimg{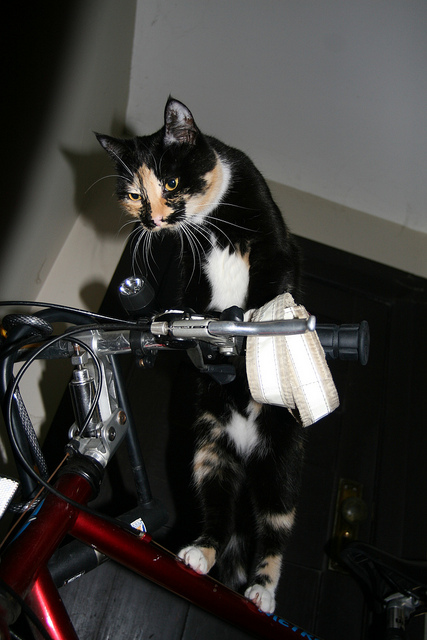}
    % \vspace{4pt}
    {\scriptsize
    \subfigheader{b}{cat $\rightarrow$ person}\\[2pt]
    \begin{tabular}{@{}l@{\hspace{0.7em}}p{0.82\linewidth}@{}}
        \caplabel{ORIG} & A cat sitting on top of a bike.\\
                        % FIXED LINE BELOW
                        & \metriclabel{E} \metricval{0.301},\; \metriclabel{S} \metricval{-0.079} \\ \addlinespace[2pt]
        \caplabel{FLIP} & A \targetword{person} sitting on top of a bike.\\
                        & \metriclabel{E} \metricval{0.210},\; \metriclabel{S} \highlight{-0.007}
    \end{tabular}
    }
\end{minipage}
\\[1.5em] % Vertical gap between rows

% ==========================
% Row 2
% ==========================

% ----- Subfigure (c) -----
\begin{minipage}[t]{0.48\linewidth}
    \centering
    \fixedheightimg{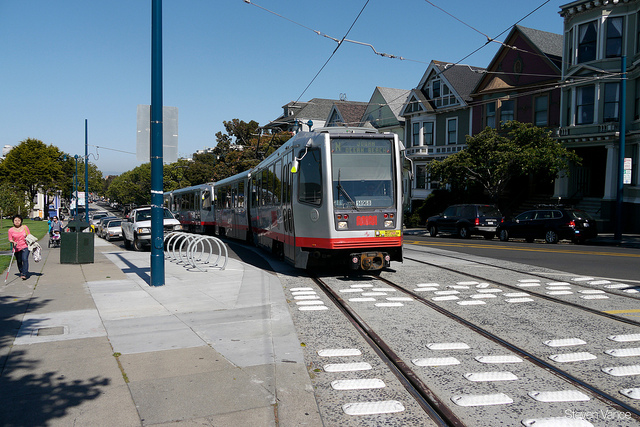}
    \vspace{-0.8cm}
    {\scriptsize
    \subfigheader{c}{train $\rightarrow$ person}\\[2pt]
    \begin{tabular}{@{}l@{\hspace{0.7em}}p{0.82\linewidth}@{}}
        \caplabel{ORIG} & A train going down the tracks in a city.\\
                        % FIXED LINE BELOW
                        & \metriclabel{E} \metricval{0.220},\; \metriclabel{S} \metricval{-0.096} \\ \addlinespace[2pt]
        \caplabel{FLIP} & A \targetword{person} going down the tracks in a city.\\
                        & \metriclabel{E} \metricval{0.195},\; \metriclabel{S} \highlight{-0.028}
    \end{tabular}
    }
\end{minipage}
&
% ----- Subfigure (d) -----
\begin{minipage}[t]{0.48\linewidth}
    \centering
    \fixedheightimg{}
    \vspace{-0.5cm}
    {\scriptsize
    \subfigheader{d}{horse $\rightarrow$ person}\\[2pt]
    \begin{tabular}{@{}l@{\hspace{0.7em}}p{0.82\linewidth}@{}}
        \caplabel{ORIG:} & a horse with reins tied up to a tree.\\
                        & \metriclabel{E} \metricval{0.267},\; \metriclabel{S} \metricval{-0.095} \\ \addlinespace[2pt]
        \caplabel{FLIP:} & a \targetword{person} with reins tied up to a tree.\\
                        & \metriclabel{E} \metricval{0.223},\; \metriclabel{S} \highlight{-0.028}
    \end{tabular}
    }
\end{minipage}
\end{tabular}
% \vspace{5pt}
\caption{Qualitative LGIP examples comparing EVA02-CLIP (E) and SigLIP (S) on object flips. In all four cases EVA assigns higher similarity to the original caption, while \textbf{SigLIP prefers the flipped caption} (scores in red), indicating a lack of semantic sensitivity to object substitutions.}
\label{fig:qualitative}
\end{figure}

Figure~\ref{fig:qualitative} highlights cases where CLIP-style models strongly prefer the human caption over a contradictory flip, whereas SigLIP assigns similar scores or prefers the flip, matching the quantitative findings.

\section{Discussion and Conclusion}
\label{sec:discussion}

We introduced \emph{Language-Guided Invariance Probing} (LGIP), a behavioral benchmark
that probes vision--language models under controlled textual perturbations while
holding the image fixed. By separating meaning-preserving paraphrases from
meaning-changing semantic flips, LGIP measures two complementary properties:
\emph{linguistic invariance} and \emph{semantic sensitivity}.

Across nine VLMs, we find these properties are not guaranteed by scale or strong
zero-shot accuracy. EVA02-CLIP and large OpenCLIP models exhibit a favorable
trade-off, combining low paraphrase-induced variance with strong rejection of
semantically inconsistent captions. In contrast, SigLIP-family models show higher
invariance error and can prefer flipped captions over human descriptions, revealing
systematic, attribute-level failures that conventional benchmarks often miss.

Beyond diagnosis, LGIP suggests concrete paths for improvement. Contrastive training
can be augmented with \emph{structured negative captions} generated by semantic flips,
and robustness can be better controlled by jointly optimizing paraphrase-consistency
and flip-discrimination losses, optionally with a strength-aware curriculum that
progressively increases semantic conflict. These diagnostics also inform downstream
behavior: low semantic sensitivity can yield linguistically plausible yet visually
incorrect rankings in image--text retrieval, while weak object-level grounding can
increase hallucination and prior-reliance in VQA. Overall, LGIP provides a compact,
model-agnostic probe that links evaluation signals to actionable training and
application-oriented insights.

\paragraph{Limitations}
Our semantic flips are restricted to a small, fixed vocabulary and single-token
substitutions, so they do not capture more compositional contradictions. Coverage also depends on whether captions contain matchable tokens, which can bias estimates toward frequent lexical patterns. 

\paragraph{Future work}
Future work includes expanding perturbations to compositional edits (e.g., negation, relations, multi-attribute changes) with larger vocabularies and automatic or human validation, improving coverage by stratifying results by token frequency and caption structure, and extending LGIP to multilingual and domain-shifted settings. Applying LGIP to instruction-tuned and generative vision--language models is also a promising direction to test whether the same invariance and sensitivity patterns hold.


\bibliographystyle{elsarticle-num-names}
\bibliography{ref}

\clearpage
\appendix
\section{Perturbation Generation Rules for LGIP}
\label{app:generation_rules}

This appendix specifies the exact procedure used to generate paraphrases and semantic
flips for LGIP, consistent with our released preprocessing pipeline. Perturbations are
generated independently for each original caption $c$ (per image--caption sample). We
distinguish meaning-preserving paraphrases (\texttt{same\_caps}) from meaning-changing
semantic flips (\texttt{diff\_caps} and type-specific fields). All sampling uses a fixed
random seed (\texttt{seed=42}).

\subsection{Paraphrase Generation (\texttt{same\_caps})}
Given an original caption $c$, we construct a candidate paraphrase set $\mathcal{P}(c)$
from two families: (i) simple style templates and (ii) advanced paraphrase rules. We
then filter, deduplicate, remove the original caption, and sample a capped set.

\paragraph{(i) Simple template paraphrases}
We instantiate the following $9$ templates:
\begin{itemize}\setlength{\itemsep}{1pt}
\item \texttt{a photo of \{c\}}
\item \texttt{an image of \{c\}}
\item \texttt{a picture of \{c\}}
\item \texttt{\{c\}}
\item \texttt{\{c\} in the scene}
\item \texttt{a scene showing \{c\}}
\item \texttt{In this image, \{c\}}
\item \texttt{In the picture, \{c\}}
\item \texttt{This image shows \{c\}}
\end{itemize}

\paragraph{(ii) Advanced paraphrase rules (\texttt{adv\_paras})}
In addition to templates, we generate advanced paraphrases using a small set of
linguistic edit rules designed to preserve the caption meaning while changing surface
form. Advanced rules are applied to the original caption $c$ to produce candidate
paraphrases; multiple rules may apply, yielding multiple candidates.

\begin{itemize}\setlength{\itemsep}{2pt}
\item \textbf{Passive voice conversion.} When an eligible active-voice pattern is detected
(e.g., a verb with an explicit direct object), we generate a passive-voice variant (e.g.,
``a person holds a cup'' $\rightarrow$ ``a cup is held by a person''). We only apply this
rule when the conversion is syntactically valid and does not drop arguments.
\item \textbf{Synonym substitution.} We replace a small subset of adjectives and verbs using a
curated synonym table. To prevent semantic drift and to avoid interfering with semantic flip
attributes, we \emph{exclude} tokens that belong to the flip vocabularies (object category,
color, and number), and we avoid replacing proper nouns or tokens that change entity identity.
\item \textbf{Syntactic restructuring.} We perform lightweight structure edits that preserve
meaning, such as swapping the order of prepositional phrases, optionally inserting discourse
markers (e.g., ``in the scene,'' ``in the image,''), or minor re-orderings that keep the same
content words and relations.
\end{itemize}

\paragraph{Simple vs. advanced labels}
For analysis, we tag a paraphrase as \textbf{simple} if it is produced by any of the
templates above (detected by the presence of template markers such as
\texttt{a photo of}, \texttt{an image of}, \texttt{a picture of}, \texttt{in this image},
\texttt{in the picture}, \texttt{this image shows} in the generated string, case-insensitive).
All other paraphrases produced by the advanced rules are tagged as \textbf{advanced}.

\paragraph{Filtering and deduplication}
We discard candidates with fewer than $5$ characters after trimming whitespace.
We then deduplicate paraphrases by exact string match, preserving the first occurrence
(order-preserving unique). Finally, we remove the exact original string $c$ from
$\mathcal{P}(c)$ if present.

\paragraph{Sampling cap}
From the remaining unique paraphrases, we sample up to $K_{\text{same}}{=}6$ paraphrases
uniformly at random without replacement (\texttt{seed=42}).

\subsection{Semantic Flip Generation (\texttt{diff\_caps} and flip-type fields)}
We generate semantic flips by performing at most one token substitution from a predefined
replacement vocabulary, using word-boundary regex matching to avoid partial-word substitutions.

\paragraph{Replacement vocabularies}
We use the following word lists (sizes in parentheses):
\begin{itemize}\setlength{\itemsep}{1pt}
\item \textbf{Color} ($11$): \texttt{\{red, blue, green, yellow, black, white, brown, gray, orange, pink, purple\}}
\item \textbf{Number} ($5$): \texttt{\{one, two, three, four, five\}}
\item \textbf{Object} ($11$): \texttt{\{dog, cat, horse, car, bus, train, person, bird, boat, bicycle, truck\}}
\end{itemize}

\paragraph{Flip rule}
For each flip type $t \in \{\texttt{color}, \texttt{number}, \texttt{object}\}$, we search
$c$ for the first occurrence of any vocabulary term using a word-boundary regex pattern.
If no match is found, no flip is generated for that type. If a match is found for a term
$w$, we uniformly sample a replacement $w' \neq w$ from the same vocabulary and substitute
only the first matched occurrence (i.e., \texttt{count=1} in regex replacement). This yields
at most one flipped caption per type.

\paragraph{Filtering and deduplication}
We discard flips that are identical to $c$ or shorter than $5$ characters after trimming
whitespace. We deduplicate flips by exact string match (order-preserving unique). If an
identical flipped string is produced by multiple types (rare), we keep the type label from
the first occurrence.

\paragraph{Sampling cap}
We sample up to $K_{\text{diff}}{=}6$ flips uniformly at random without replacement
(\texttt{seed=42}). In practice, because the rule produces at most one flip per type, the
maximum number of flips per caption is $\le 3$.

\paragraph{Output fields}
For each original caption index $i$, we store:
\begin{itemize}\setlength{\itemsep}{1pt}
\item \texttt{same\_caps[i]}: list of sampled paraphrases,
\item \texttt{diff\_caps[i]}: list of flip strings (or structured items),
\item \texttt{diff\_caps\_color[i]}, \texttt{diff\_caps\_number[i]}, \texttt{diff\_caps\_object[i]}:
lists of flips stratified by attribute type.
\end{itemize}
If \texttt{diff\_caps[i]} stores structured items, each item contains the flipped caption
text and a \texttt{flip\_type} label (e.g., \texttt{color}, \texttt{number}, \texttt{object}).

\subsection{Determinism and reproducibility}
All sampling steps use a fixed random seed (\texttt{seed=42}). Deduplication is deterministic
and order-preserving. These rules fully determine the perturbation sets used in our experiments.

% \section{Example Appendix Section}





% %% For numbered reference style

%   \textit{\LaTeX: a document preparation system},
%   Addison Wesley, Massachusetts,

% \end{thebibliography}
\end{document}